\documentclass{article} 
\pdfoutput=1
\usepackage{iclr2020_conference,times}
\usepackage{color}

\usepackage{amsmath,amsfonts,bm}

\def\eqref#1{equation~\ref{#1}}
\def\Eqref#1{Equation~\ref{#1}}

\def\1{\bm{1}}

\DeclareMathAlphabet{\mathsfit}{\encodingdefault}{\sfdefault}{m}{sl}
\SetMathAlphabet{\mathsfit}{bold}{\encodingdefault}{\sfdefault}{bx}{n}

\usepackage{graphicx}
\usepackage{hyperref}
\usepackage{subfig}
\usepackage{wasysym}
\usepackage{url}
\usepackage{soul}
\definecolor{new_green}{HTML}{D0F278}

\title{Train, Learn, Expand, Repeat.}

\author{Abhijeet Parida\textsuperscript{1}\thanks{These authors contributed equally to this work.}, Aadhithya Sankar\textsuperscript{1,2,*}, Rami Eisawy\textsuperscript{1,2}, Tom Finck\textsuperscript{2}, Benedikt Wiestler\textsuperscript{2},\\\textbf{ Franz Pfister\textsuperscript{1,3} \& Julia Moosbauer\textsuperscript{1,3}}  \\
\textsuperscript{1}deepc GmbH\\
\textsuperscript{2}Technische Universit\"at M\"unchen \\
\textsuperscript{3}Ludwig‑-Maximilians--Universit\"at M\"unchen \\
\texttt{\{abhijeet.parida,aadhithya.sankar,rami.eisawy\}@deepc.ai},\\ \texttt{\{tom.finck, b.wiestler\}@tum.de},\\ \texttt{\{franz.pfister,julia.moosbaer\}@deepc.ai} }

\iclrfinalcopy 
\begin{document}
\maketitle
\begin{abstract}

High-quality labeled data is essential to successfully train supervised machine learning models. Although a large amount of unlabeled data is present in the medical domain, labeling poses a major challenge: medical professionals who can expertly label the data are a scarce and expensive resource. Making matters worse, voxel-wise delineation of data (e.g. for segmentation tasks) is tedious and suffers from high inter-rater variance, thus dramatically limiting available training data.

We propose a recursive training strategy to perform the task of semantic segmentation given only very few training samples with pixel-level annotations. We expand on this small training set having cheaper image-level annotations using a recursive training strategy. 
We apply this technique on the segmentation of intracranial hemorrhage (ICH) in CT (computed tomography) scans of the brain, where typically few annotated data is available.

\end{abstract}
\section{Introduction}

Deep artificial neural networks like convolutional neural networks (CNNs) are currently the state-of-the-art for semantic segmentation of both natural and medical images \citep{dolz}. This is achieved by leveraging the information from large, well-labeled datasets with ground truth annotations. However, in the medical domain, large delineated dataset of high quality are difficult to generate. \citet{hans} explain that while the annotation of medical images is done by highly specialized physicians, the resulting segmentation is still very prone to inter/intra-observer variability. Algorithms that can learn from unlabeled or weakly labeled training data are essential in the medical domain to leverage the vast amount of already available unlabeled data.

Datasets equipped with image labels are much more readily available and in their absence are easier and faster to create \citep{COCO}. As such, we propose a recursive method that is able to transfer knowledge obtained from a small, fully-supervised segmented dataset to obtain segmentation for a larger weakly-supervised dataset. 

Our main contributions are: 1) We present a semi-supervised recursive learning strategy that transfers knowledge gained from a small segmented dataset to a larger weakly labeled dataset. 2) We apply the algorithm to the task of semantic segmentation of intracranial hemorrhage in brain CT scans, and show that the results on several datasets. 


\section{Related Work}

Accurate annotation of medical imaging data is pivotal for the successful training of supervised algorithms. It is very difficult to find datasets which are both large enough for training robust models, and where the annotation quality is high. To mitigate this problem, several strategies have been explored to reduce the need for such datasets.

One strategy to reduce the amount of training data needed is \emph{transfer learning}: \citet{pan2010survey} describe transfer learning as a technique which relays knowledge acquired from a domain with rich data availability to a domain with low data availability. Transfer learning techniques can be applied both across domains and tasks. 


Another approach is \emph{semi-supervised learning}, where algorithms are trained on a dataset that contains a small amount of labeled data and a large portion of unlabeled examples. One popular semi-supervised technique is self-training \citep{SSLSurvey}. The idea in self-learning is to propagate predictions from the small labeled data to the large unlabeled data and
subsequently use the newly created labeled set for training. \citet{zhu2009introduction} advocate that this approach assumes that the method's high confidence predictions are correct and can be used further.
As seen in \citet{su2015interactive}, an active verification step in which a human is queried to verify some of the labels can be deployed to avoid error propagation. 

In situations where an imprecise or inaccurate annotation of the data is much cheaper to obtain compared to an accurate annotation, \emph{weakly-supervised learning} can be taken into consideration. \citet{simple} argue that a large number of noisy annotations should convey enough information about the task to be performed with reasonable accuracy while reducing the burden on the data annotator. The aim is thus to replace time-consuming annotation procedures, yielding the way for potentially labeling larger datasets. \citet{boxnet} have shown it is possible to successfully learn a segmentation task in the medical domain by using axis-aligned bounding boxes, derived from 6 points. Instead of full, pixel-wise segmentation labels, less time consuming labelling techniques include bounding boxes \citep{simple,boxnet}, scribbles \citep{scribblesup}, or points \citep{,rakelly}. 

\section{Proposed Method}

\begin{figure}[ht]
    \centering
    \includegraphics[width=400px, height=175px]{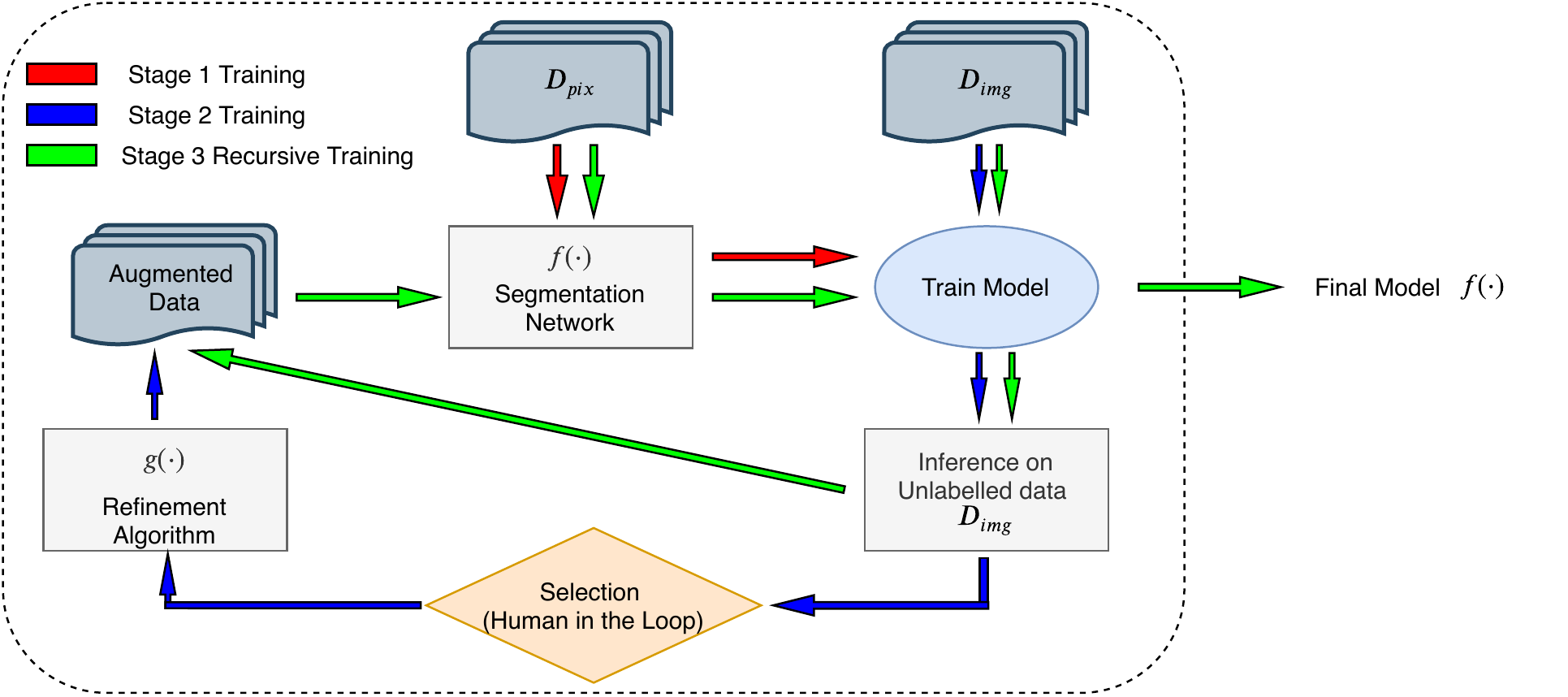}
    \caption{The flowchart depicts the schema for the proposed three-stage training strategy. The first stage, marked in red, shows the supervised learning on a small, but fully segmented dataset $\mathcal{D}_{\mathrm{pix}}$. The second stage, marked in blue, shows the samples with good segmentation that are actively selected to augment the dataset from the fully supervised approach. Finally, in the third stage, shown in green, the training continues recursively until the stopping criteria are reached. }
    \label{fig:Proposed training routine}
\end{figure}

To perform the task of semantic segmentation in data regimes where only few segmented data is available, we propose a new multi-stage semi-supervised training strategy as shown in Figure \ref{fig:Proposed training routine}, supported by a human-in-the-loop. The idea combines the ability of CNNs to efficiently learn from few training samples with a recursive de-noising training strategy similar to that seen in \citet{simple}, and thus yields segmentations of high quality, while only requiring few annotated input samples. 

We consider a semi-supervised setting, where $\mathcal{D}_{\mathrm{pix}}$ denotes an image dataset equipped with precise pixel-level segmentations, i.e. for each image $x \in \mathcal{D}_{\mathrm{pix}}$ segmentations $y_{\mathrm{pix}}$ for the respective classes $\{1, ..., K + 1\}$ are available. For $\mathcal{D}_\mathrm{img}$, however, only image-level class-labels $y_{\mathrm{img}}\in \{1, ..., K + 1\}$ are available. Since full segmentations are more expensive to generate, we assume that $\mathcal{D}_\mathrm{img}$ is considerably larger than $\mathcal{D}_\mathrm{pix}$. 

In the first stage of training, a naive function approximator $f$ is trained on $\mathcal{D}_{\mathrm{pix}}$ to segment an image $x$ of width $W$ and height $H$ with respect to $K + 1$ classes, i.e.\ $f(x) \in \{0, 1\}^{W, H, K + 1}$. This naive baseline is a deep neural network which has been trained to overfit on $\mathcal{D}_{pix}$.
The learning is done by minimizing the multi-class cross entropy loss
\begin{equation}
\mathcal{L}\left(x, y_{\mathrm{pix}}\right) =\frac{-1}{H \cdot W}\sum_j \sum_k y_{\mathrm{pix}, k}^{(j)} \log f_k\left(x^{( j)}\right)
\label{Eqn:cross-entropy}
\end{equation}
 with $y_{\mathrm{pix}, k}^{(j)}$ denoting the $j$-th pixel of the segmentation belonging to class $k$. 
  
In the second stage of training, the model $f$ is used to create segmentations for $x \in \mathcal{D}_\mathrm{img}$, which serves as an cheap but inaccurate approximation to the unknown ground truth segmentation of the image, i.e., a weak label is generated. In the selection step shown in Figure \ref{fig:Proposed training routine} a human expert identifies those segmentations with a low number false positives. The extent of the true positives is not taken into consideration at all. The expert is not allowed to make any changes to the annotation. The selected subset of samples is then used to augment the original dataset $\mathcal{D}_\mathrm{pix}$.


Before the selected samples are used to augment the original dataset $D_\mathrm{img}$, a step of post-processing may applied to the generated weak labels to refine them further. If it can be assumed that the generated labels are larger than the true but unkown ground-truth segmentations, the segmentation algorithm $g$ may be chosen such that it reduces the false positives. If, in contrast, it can be assumed that the generated weak labels are smaller than the unknown ground truth, $g$ can be chosen such that it increases the true positives. Another criterion that $g$ could enforce is the \emph{objectness} \citep{simple}, to make disjoint objects continuous and to force existing boundaries in $x$ on the respective segmentations.


In the final stage, $f$ is recursively re-trained by minimizing \Eqref{Eqn:cross-entropy}. The loss is minimized between the segmentation generated by $f$ in the previous recursion, as ground truth, with the current recursion's prediction for $x \in \mathcal{D}_\mathrm{img}$. After each recursion, images which are newly segmented from $\mathcal{D}_\mathrm{img}$, are added to the training data. For the images $x \in \mathcal{D}_\mathrm{pix}$, where the ground truth, $y_\mathrm{pix}$, is available, the objective is to minimize the \Eqref{Eqn:cross-entropy} and the dice loss \citep{dice} as a regularization. The main idea exploited by the recursion is that the noisy segmentation generated will lead to better generalization on the inputs and generate robust predictions\citep{simple}.The dice loss will allow for controlled growth/shrinkage of the segmentation learned by $f$. This recursive process continues until the network no longer expands to new data.

\section{Semantic Segmentation of Intracranial Hemorrhage}

We choose the task of ICH segmentation in brain CT scans for validating the proposed approach. It satisfies all the criteria necessary for the application of the methods presented in this paper: (1) it is difficult and time consuming to create the segmentation labels, (2) no large segmented datasets are publicly available, and (3) large datasets with image-level class details are available.

\subsection{Datasets}
The datasets that we make use of are the PhysioNet dataset \citep{physionet}, which corresponds to the $\mathcal{D}_\mathrm{pix}$ with the full pixel-level annotation, and the dataset provided in the RSNA Intracranial Hemorrhage Detection challenge\footnote{\url{https://www.kaggle.com/c/rsna-intracranial-hemorrhage-detection/data}}. From the RSNA dataset we sample images with only one class present to form our $\mathcal{D}_\mathrm{img}$ subset. Apart from these we make use of an in-house dataset of brain CT scans with ICH\footnote{We have obtained positive ethical vote from the ethical committee of Technical University Munich to use the data for research purposes (344/19-SR).} and CQ 500 dataset \citep{cq500} for benchmarking the performance of the recursive strategy. Each of the dataset consists of $K = 5$ bleed classes, whose distribution can be seen in Table \ref{tab:dataset} in the Appendix \ref{chap:dataset}.

\subsection{Setup and Experiments}

For the ICH segmentation task, we choose UNet \citep{unet} as the segmentation network $f$ and the Felzenszwalb-Huttenlocher \citep{felzen} algorithm as the approximator $g$ to refine the labels produced by the UNet.

Initially, the UNet was trained on the Physionet dataset for 120 epochs before running inference on the RSNA dataset. We then manually selected segmentations from the predictions to use as ground truth for the recursive training of the UNet. In each recursion, the UNet was trained for three epochs, with the results from the recursion used as the ground truth for the next recursion.

To test the models, a radiologist was given the task to label 20 pathological volumes from the CQ 500 \citep{cq500} dataset. Additionally, the models were tested on an in-house dataset consisting of various ICHs. We evaluate the model's performance using the \emph{Dice Coefficient} and \emph{Intersection over Union Scores} \citep{dice} on both datasets. 

\subsection{Results}

Test results performed on the two unseen datasets are provided in Table \ref{tab:results}. We observe that the recursive strategy provides a significant improvement in performance on our internal dataset. There is a slight decrease in performance on the fully supervised dataset, which we assume to be due to the increased generalizing capacity of $f$. Put differently, the minor details specific to the Physionet dataset have been ignored but the large trends are still maintained.

\begin{table}[ht]
\caption{The segmentation results after performing the recursive training strategy across datasets. }
\begin{center}
\begin{tabular}{llcccc}
\multicolumn{1}{l}{\bf Dataset} &
\multicolumn{1}{l}{\bf Model} &
\multicolumn{1}{c}{\bf Dice}& \multicolumn{1}{c}{\bf IoU}&
\multicolumn{1}{c}{\bf Precision}& \multicolumn{1}{c}{\bf Recall}
\\ \hline \hline \\

In-House& Before&$0.406( 0.424)$&$0.322(0.283)$& $ 0.520(.683) $&$0.380(0.551)$\\
Dataset& Recursion &$\pm 0.127$&$\pm 0.091$& $\pm0.160$&$\pm0.112$\\ \cline{2-6}
& After &$0.485 (0.616)$&$0.402( 0.445)$& $ 0.575 (0.829)$& $0.461 (0.500)$\\
&Recursion&$\pm 0.140$&$\pm 0.112$& $ \pm 0.163$& $\pm 0.147$\\
\hline
Median Change&&{\color{green}\UParrow} $0.192 $ &{\color{green}\UParrow} $ 0.162$ & {\color{green}\UParrow} $0.146 $& {\color{red}\DOWNarrow} $0.051 $\\
\hline\hline

CQ 500& Before&$0.417( 0.478) $&$0.304 (0.315)$&$0.561 (0.676)$ &$0.362 (0.358)$\\

& Recursion&$\pm0.077$&$\pm0.053$&$ \pm0.128$ &$\pm 0.070 $\\ \cline{2-6}

& After&$0.446(0.553) $&$0.343(0.385)$&$0.629(0.766) $&$0.390( 0.415)$\\
&  Recursion&$\pm   0.105$&$\pm 0.072$&$ \pm0.147$&$\pm 0.100$\\
\hline
Median Change&&{\color{green}\UParrow} $0.075 $ &{\color{green}\UParrow} $0.070 $ & {\color{green}\UParrow} $0.090 $& {\color{green}\UParrow} $ 0.057$\\
\hline\hline

Physionet& Before&$0.611 (0.754)$&$0.534 (0.605)$&$0.734 (0.893)$ &$0.586 ( 0.758)$\\
& Recursion&$\pm 0.135$&$\pm 0.130$&$ \pm 0.125$ &$\pm 0.147$\\ \cline{2-6}
& After &$0.610 (0.733) $&$0.533 (0.570) $&$0.723 (0.887) $&$0.575 (0.670) $\\
&Recursion&$ \pm 0.132$&$ \pm  0.131$&$\pm 0.131$&$\pm 0.144$\\
\hline
Median Change&&{\color{red}\DOWNarrow} $ 0.021$ &{\color{red}\DOWNarrow} $0.035 $ & {\color{red}\DOWNarrow} $0.006 $& {\color{red}\DOWNarrow} $0.088 $\\
\hline\hline
\end{tabular}
\end{center}\label{tab:results}
\end{table}

\begin{figure}[h]
  \centering
{
       \includegraphics[height=0.20\textheight,width=0.48\textwidth]{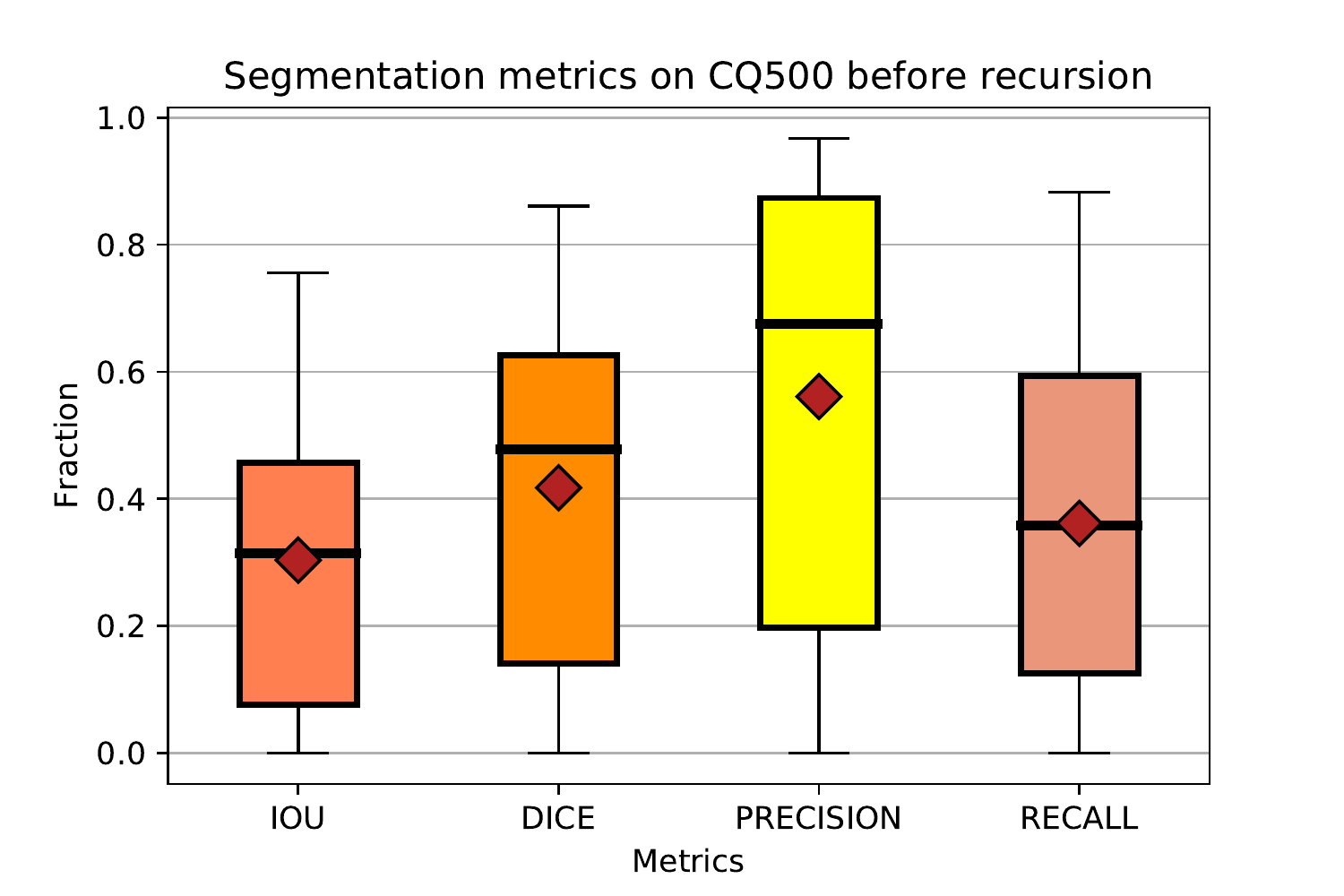}
     }
     \hfill
{
       \includegraphics[height=0.20\textheight,width=0.48\textwidth]{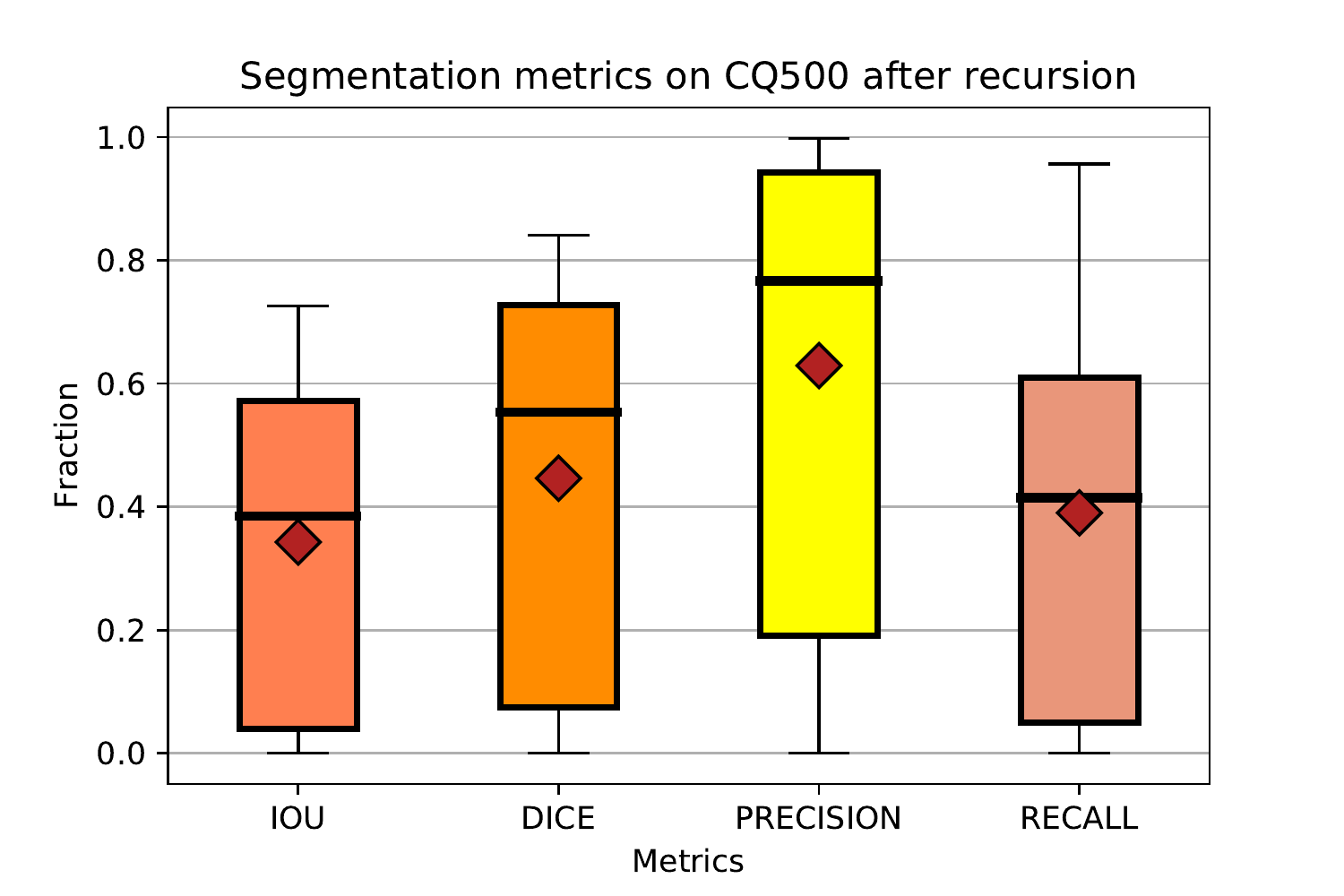}
     }
  \caption{Boxplot for segmentation metrics on CQ 500. The complete boxplot for all the datasets is available in Appendix \ref{chap:result} Fig.\ref{fig:boxplot} }
  \label{fig:boxplot_CQ}
\end{figure}

The CQ 500 patient performance details are shown in the boxplots in Figure \ref{fig:boxplot_CQ}. The increase in performance after applying the recursive training strategy is lower compared to that on our internal dataset, due to the noisy nature of the scans in the dataset. We expect that training entirely on noisy input images could help to improve the performance globally and would allow further model generizability.

Overall, there are two kinds of trends seen in the performance of the model $f$ before and after applying the recursive training strategy. First, $f$ can identify new regions that would not have been segmented without recursion (Figure \ref{fig:seg_CQ500_new} in Appendix \ref{chap:visual}). Second, with recursion, the segmentation quality of identified regions is improved (Figure \ref{fig:seg_CQ500_main} and Figures \ref{fig:zoom_fig}, \ref{fig:seg_CQ500_good} in Appendix \ref{chap:visual}). 


\begin{figure}[h]
  \centering
      \includegraphics[ height=0.28\textheight,width=1\textwidth]{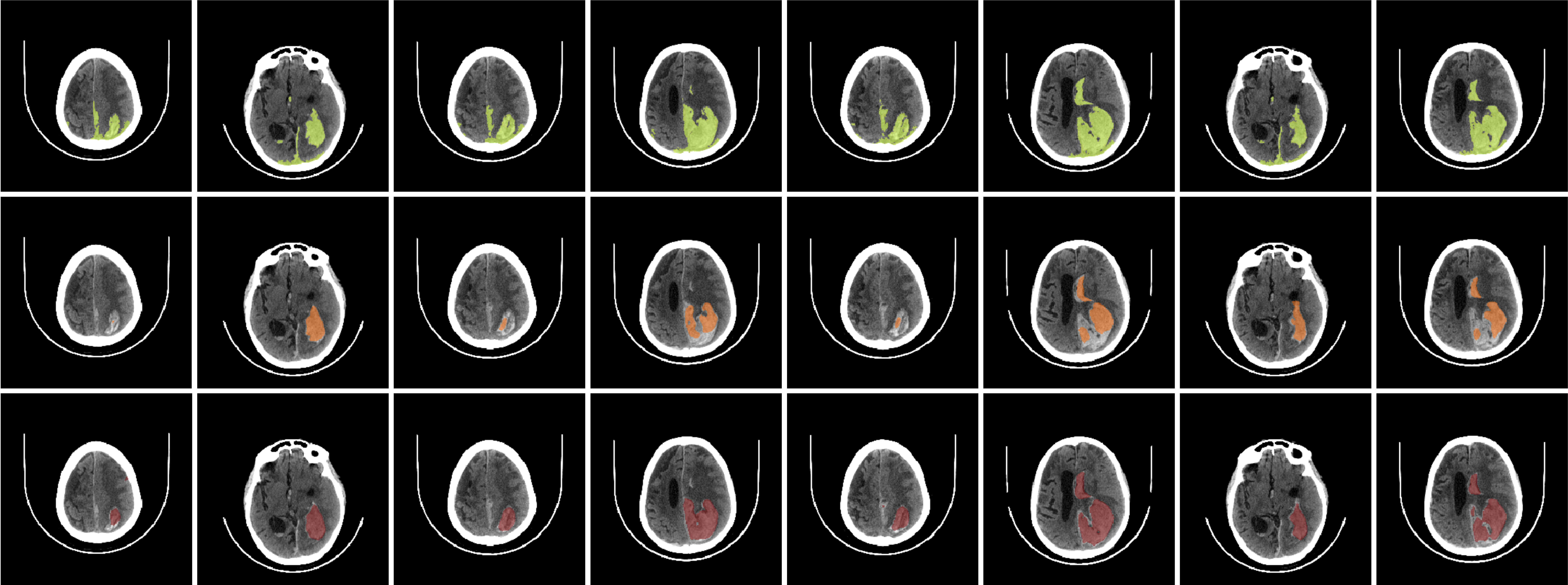}
  \caption{Predicted segmentation masks from the model on the CQ 500. The green regions($1^{st}\ row$) correspond to the {\color{black}ground truth}, the orange regions($2^{nd}$ row) show segmentations {\color{black} without the recursive strategy} and the red regions ($3^{rd}$ row) show segmentations {\color{black} after recursion}. More samples from CQ 500 can be seen in Appendix \ref{chap:visual}. }.
  \label{fig:seg_CQ500_main}
\end{figure}

\section{Conclusion}

We propose a new training scheme for segmentation tasks in situations where only little ground truth data is available. Our results indicate that starting with a small labeled dataset and recursively expanding into a new dataset without pixel-level annotations may improve the networks ability generalize across different datasets. We demonstrated this with the task of Intracranial Hemorrhage Segmentation using the PhysioNet dataset as initial labeled dataset and the RSNA ICH dataset as the unlabeled expansion dataset. 

Especially in the medical sector there are various weakly semi-supervised data scenarios where our proposed training scheme can be applied successfully. The recursive training scheme, which combines semi-supervised and transfer learning with a human-in-the-loop approach, is so general that we think other medical use cases and tasks other then ICH segmentation could benefit from it. 

\clearpage

\bibliography{references}

\begin{thebibliography}{16}
\providecommand{\natexlab}[1]{#1}
\providecommand{\url}[1]{\texttt{#1}}
\expandafter\ifx\csname urlstyle\endcsname\relax
  \providecommand{\doi}[1]{doi: #1}\else
  \providecommand{\doi}{doi: \begingroup \urlstyle{rm}\Url}\fi

\bibitem[B{\o} et~al.(2017)B{\o}, Solheim, Jakola, Kvistad, Reinertsen, and
  Berntsen]{hans}
Hans~Kristian B{\o}, Ole Solheim, Asgeir~Store Jakola, Kjell-Arne Kvistad,
  Ingerid Reinertsen, and Erik~Magnus Berntsen.
\newblock Intra-rater variability in low-grade glioma segmentation.
\newblock \emph{Journal of Neuro-Oncology}, 131\penalty0 (2):\penalty0
  393--402, Jan 2017.
\newblock ISSN 1573-7373.
\newblock \doi{10.1007/s11060-016-2312-9}.
\newblock URL \url{https://doi.org/10.1007/s11060-016-2312-9}.

\bibitem[Cheplygina et~al.(2018)Cheplygina, de~Bruijne, and Pluim]{SSLSurvey}
Veronika Cheplygina, Marleen de~Bruijne, and Josien P.~W. Pluim.
\newblock Not-so-supervised: a survey of semi-supervised, multi-instance, and
  transfer learning in medical image analysis.
\newblock \emph{CoRR}, abs/1804.06353, 2018.
\newblock URL \url{http://arxiv.org/abs/1804.06353}.

\bibitem[Chilamkurthy et~al.(2018)Chilamkurthy, Ghosh, Tanamala, Biviji,
  Campeau, Venugopal, Mahajan, Rao, and Warier]{cq500}
Sasank Chilamkurthy, Rohit Ghosh, Swetha Tanamala, Mustafa Biviji, Norbert~G.
  Campeau, Vasantha~Kumar Venugopal, Vidur Mahajan, Pooja Rao, and Prashant
  Warier.
\newblock Development and validation of deep learning algorithms for detection
  of critical findings in head {CT} scans.
\newblock \emph{CoRR}, abs/1803.05854, 2018.
\newblock URL \url{http://arxiv.org/abs/1803.05854}.

\bibitem[Dolz et~al.(2018)Dolz, Gopinath, Yuan, Lombaert, Desrosiers, and
  Ayed]{dolz}
Jose Dolz, Karthik Gopinath, Jing Yuan, Herve Lombaert, Christian Desrosiers,
  and Ismail~Ben Ayed.
\newblock Hyperdense-net: a hyper-densely connected cnn for multi-modal image
  segmentation.
\newblock \emph{IEEE transactions on medical imaging}, 38\penalty0
  (5):\penalty0 1116--1126, 2018.

\bibitem[Felzenszwalb \& Huttenlocher(2004)Felzenszwalb and
  Huttenlocher]{felzen}
Pedro~F Felzenszwalb and Daniel~P Huttenlocher.
\newblock Efficient graph-based image segmentation.
\newblock \emph{International journal of computer vision}, 59\penalty0
  (2):\penalty0 167--181, 2004.

\bibitem[{Hssayeni} et~al.(2019){Hssayeni}, {S.}, {Croock}, {Ph.}, {Al-Ani},
  {Ph.}, {Falah Al-khafaji}, {D.}, {Yahya}, {D.}, {Ghoraani}, and
  {D}]{physionet}
Murtadha~D. {Hssayeni}, M.~{S.}, Muayad~S. {Croock}, D.~{Ph.}, Aymen {Al-Ani},
  D.~{Ph.}, Hassan {Falah Al-khafaji}, M.~{D.}, Zakaria~A. {Yahya}, M.~{D.},
  Behnaz {Ghoraani}, and Ph. {D}.
\newblock {Intracranial Hemorrhage Segmentation Using Deep Convolutional
  Model}.
\newblock \emph{arXiv e-prints}, art. arXiv:1910.08643, Oct 2019.

\bibitem[{Jimenez-del-Toro} et~al.(2016){Jimenez-del-Toro}, {Müller}, {Krenn},
  {Gruenberg}, {Taha}, {Winterstein}, {Eggel}, {Weber}, {Dicente Cid}, {Gass},
  {Heinrich}, {Jia}, {Kahl}, {Kechichian}, {Mai}, {Spanier}, {Vincent}, {Wang},
  {Wyeth}, and {Hanbury}]{dice}
O.~{Jimenez-del-Toro}, H.~{Müller}, M.~{Krenn}, K.~{Gruenberg}, A.~A. {Taha},
  M.~{Winterstein}, I.~{Eggel}, {Weber}, Y.~{Dicente Cid}, T.~{Gass},
  M.~{Heinrich}, F.~{Jia}, F.~{Kahl}, R.~{Kechichian}, D.~{Mai}, A.~B.
  {Spanier}, G.~{Vincent}, C.~{Wang}, D.~{Wyeth}, and A.~{Hanbury}.
\newblock Cloud-based evaluation of anatomical structure segmentation and
  landmark detection algorithms: Visceral anatomy benchmarks.
\newblock \emph{IEEE Transactions on Medical Imaging}, 35\penalty0
  (11):\penalty0 2459--2475, Nov 2016.
\newblock ISSN 0278-0062.
\newblock \doi{10.1109/TMI.2016.2578680}.

\bibitem[Khoreva et~al.(2017)Khoreva, Benenson, Hosang, Hein, and
  Schiele]{simple}
Anna Khoreva, Rodrigo Benenson, Jan Hosang, Matthias Hein, and Bernt Schiele.
\newblock Simple does it: Weakly supervised instance and semantic segmentation.
\newblock In \emph{Proceedings of the IEEE conference on computer vision and
  pattern recognition}, pp.\  876--885, 2017.

\bibitem[Lin et~al.(2016)Lin, Dai, Jia, He, and Sun]{scribblesup}
Di~Lin, Jifeng Dai, Jiaya Jia, Kaiming He, and Jian Sun.
\newblock Scribblesup: Scribble-supervised convolutional networks for semantic
  segmentation.
\newblock In \emph{Proceedings of the IEEE Conference on Computer Vision and
  Pattern Recognition}, pp.\  3159--3167, 2016.

\bibitem[Lin et~al.(2014)Lin, Maire, Belongie, Bourdev, Girshick, Hays, Perona,
  Ramanan, Doll{\'{a}}r, and Zitnick]{COCO}
Tsung{-}Yi Lin, Michael Maire, Serge~J. Belongie, Lubomir~D. Bourdev, Ross~B.
  Girshick, James Hays, Pietro Perona, Deva Ramanan, Piotr Doll{\'{a}}r, and
  C.~Lawrence Zitnick.
\newblock Microsoft {COCO:} common objects in context.
\newblock 2014.
\newblock URL \url{http://arxiv.org/abs/1405.0312}.

\bibitem[Pan \& Yang(2010)Pan and Yang]{pan2010survey}
SJ~Pan and Q~Yang.
\newblock A survey on transfer learning. ieee transaction on knowledge
  discovery and data engineering, 22 (10), 2010.

\bibitem[Rakelly et~al.(2018)Rakelly, Shelhamer, Darrell, Efros, and
  Levine]{rakelly}
Kate Rakelly, Evan Shelhamer, Trevor Darrell, Alexei~A Efros, and Sergey
  Levine.
\newblock Meta-learning to guide segmentation.
\newblock 2018.

\bibitem[Ronneberger et~al.(2015)Ronneberger, Fischer, and Brox]{unet}
Olaf Ronneberger, Philipp Fischer, and Thomas Brox.
\newblock U-net: Convolutional networks for biomedical image segmentation.
\newblock In \emph{International Conference on Medical image computing and
  computer-assisted intervention}, pp.\  234--241. Springer, 2015.

\bibitem[Su et~al.(2015)Su, Yin, Huh, Kanade, and Zhu]{su2015interactive}
Hang Su, Zhaozheng Yin, Seungil Huh, Takeo Kanade, and Jun Zhu.
\newblock Interactive cell segmentation based on active and semi-supervised
  learning.
\newblock \emph{IEEE transactions on medical imaging}, 35\penalty0
  (3):\penalty0 762--777, 2015.

\bibitem[Yang et~al.(2018)Yang, Zhang, Zhao, Zheng, Liang, Ying, Ahuja, and
  Chen]{boxnet}
Lin Yang, Yizhe Zhang, Zhuo Zhao, Hao Zheng, Peixian Liang, Michael~TC Ying,
  Anil~T Ahuja, and Danny~Z Chen.
\newblock Boxnet: Deep learning based biomedical image segmentation using boxes
  only annotation.
\newblock \emph{arXiv preprint arXiv:1806.00593}, 2018.

\bibitem[Zhu \& Goldberg(2009)Zhu and Goldberg]{zhu2009introduction}
Xiaojin Zhu and Andrew~B Goldberg.
\newblock Introduction to semi-supervised learning.
\newblock \emph{Synthesis lectures on artificial intelligence and machine
  learning}, 3\penalty0 (1):\penalty0 1--130, 2009.

\end{thebibliography}
\bibliographystyle{iclr2020_conference}
\newpage 
\appendix
\section{Appendix}
\subsection*{Dataset Distribution}\label{chap:dataset}
PhysioNet is a small dataset with full pixel-level segmentations for each of the bleed classes. RSNA ICH is a very large dataset with image-level labels for the bleed classes in the images. We choose those training slices where only one type of bleeding is apparent. We then subsample the dataset to have equal occurence of each class.   
For the CQ 500 we consider a random subselection of 20 patients, which have been manually annotated by a neuro-radiologist.
\begin{table}[ht]
\caption{The distribution of number of slices of CT scans in each of the dataset that was used for all the experiments}
\label{sample-table}
\begin{center}
\begin{tabular}{lcccc}
\multicolumn{1}{c}{\bf Type of Bleed ($K+1$)} &
\multicolumn{1}{c}{\bf PhysioNet}  &\multicolumn{1}{c}{\bf RSNA}
&\multicolumn{1}{c}{\bf CQ 500}
&\multicolumn{1}{c}{\bf In House Datset}\\\hline \hline\\

Epidural &$173$ &$1497$&-&-\\
Intraparenchymal &$60$ &$1497$&-&-\\
Intraventricular &$13$ &$1497$&-&-\\
Subarchnoid &$16$ &$1497$&-&-\\
Subdural &$56$ &$1497$&-&-\\
\hline \hline \\
Bleed Slices&$318$&$7485$&$1025$&$1016$\\
No Bleed Slices&$75$&$300$&$2276$&$2087$\\
\hline \hline \\
\textbf{Total Slices}&$393$&$7785$&$3301$&$3103$\\
\hline \hline \\

\end{tabular}
\end{center}\label{tab:dataset}
\end{table}

The 20 patients manually annotated by the neuro-radiologist are-\\
\{\ $CT\_107,\ CT\_129,\ CT\_139,\ CT\_149, \ CT\_154,\ CT\_181,\ CT\_196,\ CT\_248, \ CT\_305,\\ \ CT\_35,  \ CT\_404,\ CT\_408, \ CT\_ 417, \ CT\_420, \ CT\_429, \ CT\_452, \ CT\_456,\ CT\_48, \\ \ CT\_4, \ CT\_ 90 $\ \}.
\subsection*{Additional Visualizations}\label{chap:visual}

\begin{figure}[!h]
   \begin{minipage}{0.33\textwidth}
     \centering
     \includegraphics[height=0.18\textheight,width=0.8\linewidth]{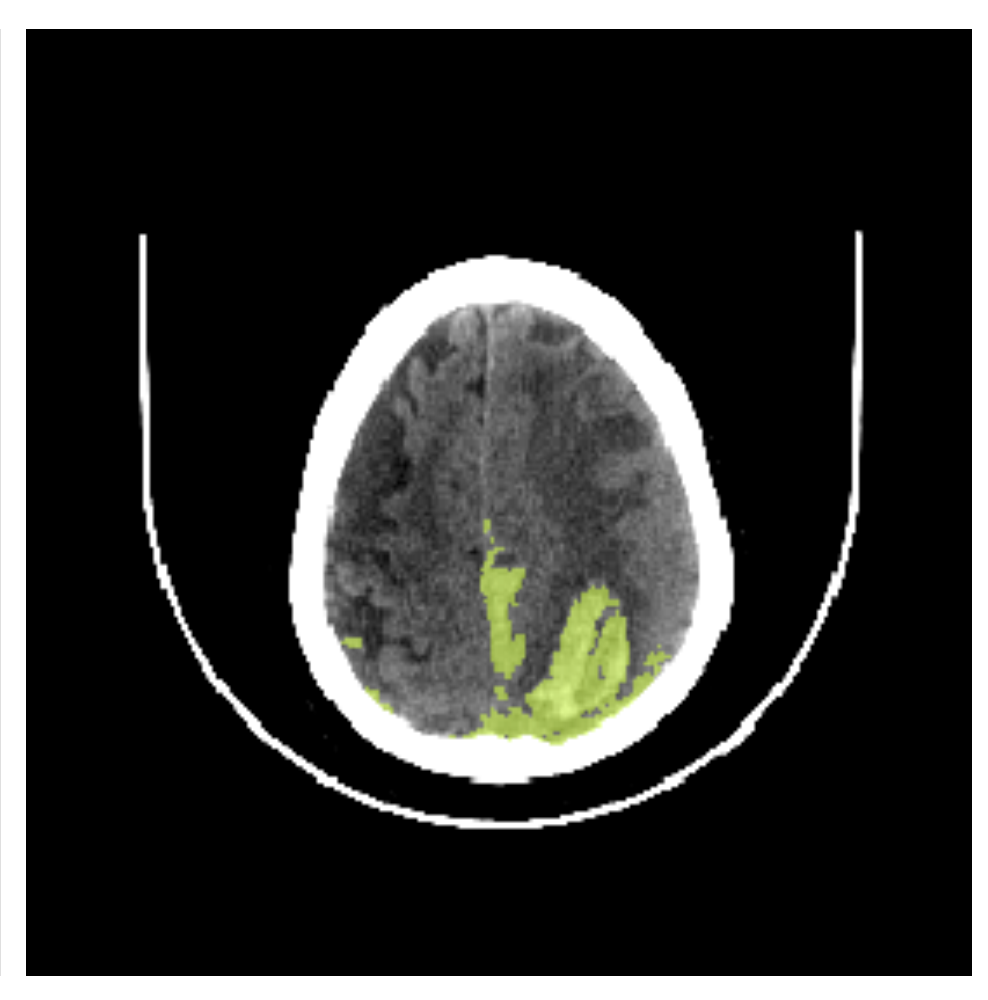}
   \end{minipage}\hfill
   \begin{minipage}{0.33\textwidth}
     \centering
     \includegraphics[height=0.18\textheight,width=0.8\linewidth]{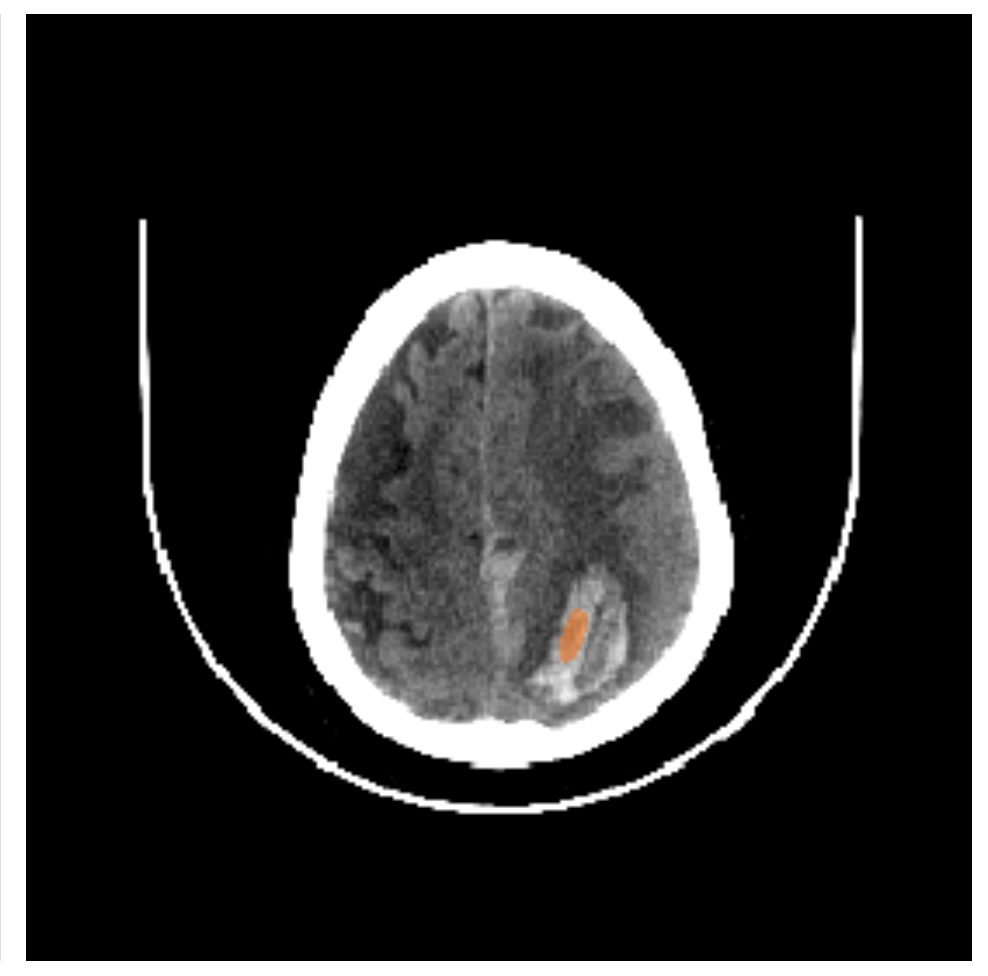}
   \end{minipage}
   \begin{minipage}{0.33\textwidth}
     \centering
     \includegraphics[height=0.18\textheight,width=0.8\linewidth]{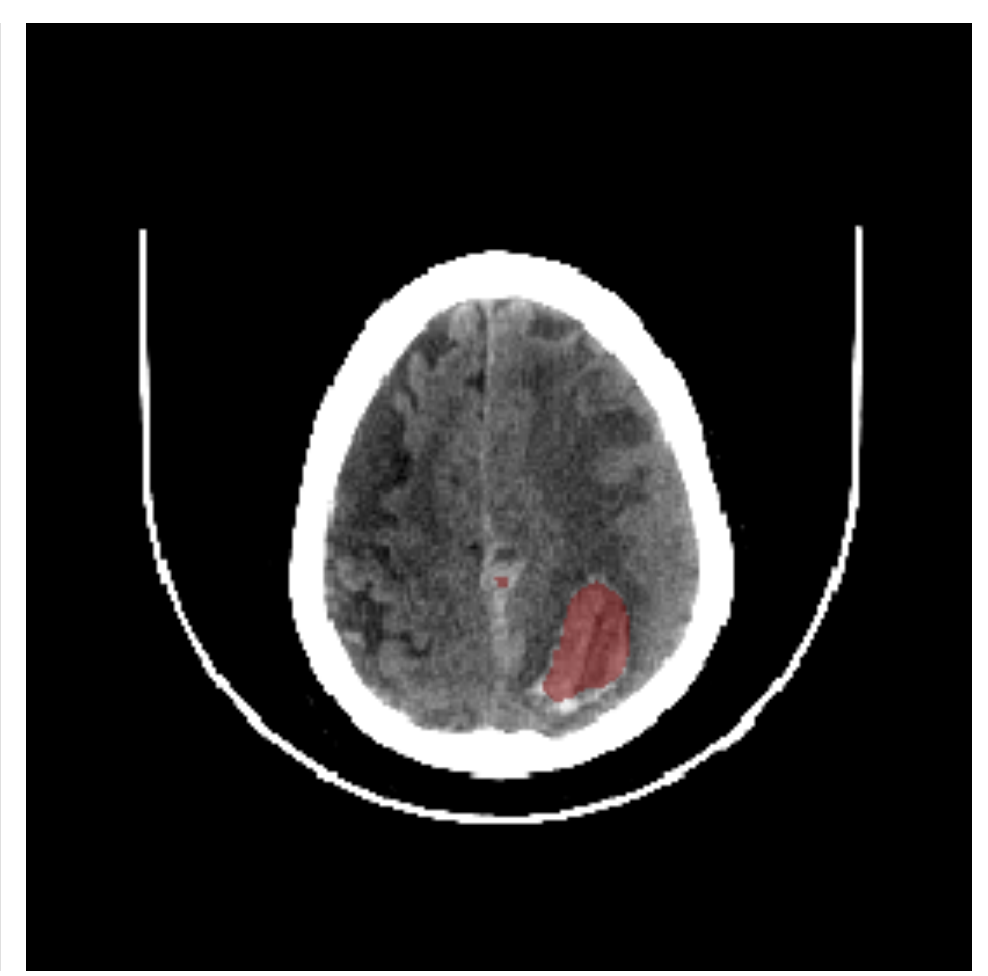}
   \end{minipage}
      \caption{Predicted results for a slice. The left image(green) is the ground truth, middle image(orange) is the output without recursive training and right output(red) with recursive training shows the model capacity in identifying new regions and expanding from seed points.}
  \label{fig:zoom_fig}
\end{figure}

\begin{figure}[ht]
  \centering
  
       \includegraphics[ height=0.30\textheight,width=1\textwidth]{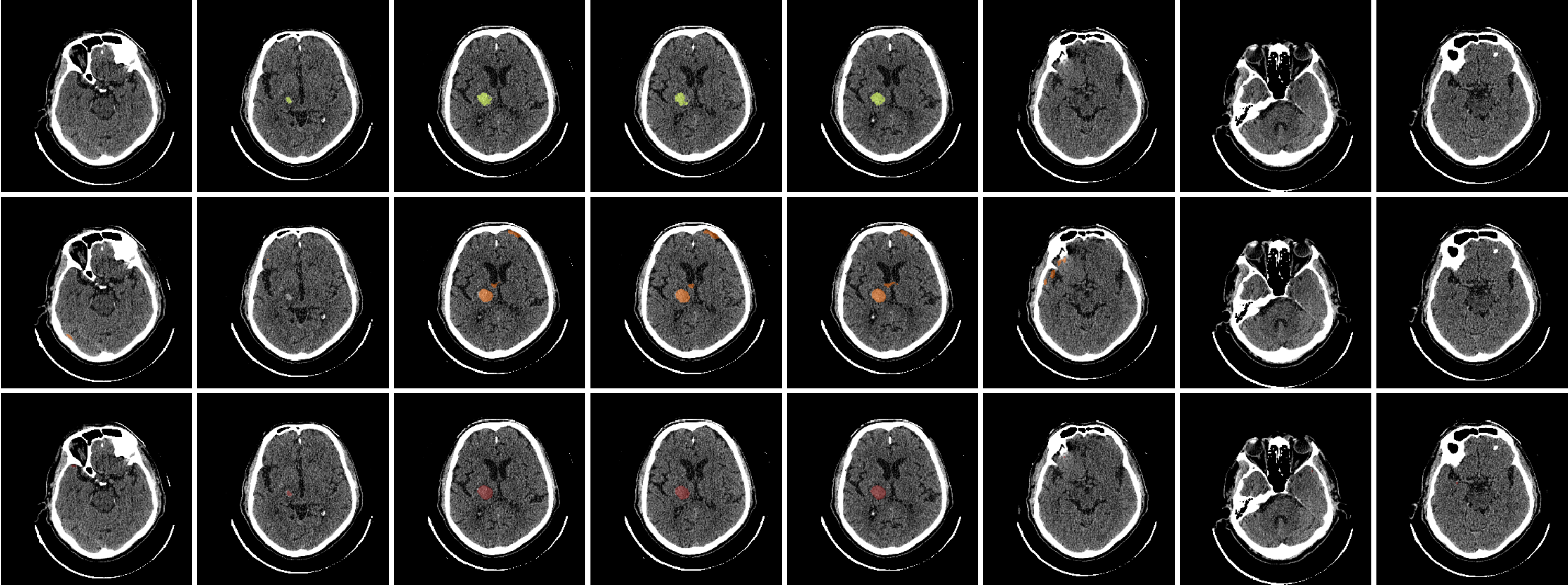}

     \vfill
     \includegraphics[ height=0.30\textheight,width=1\textwidth]{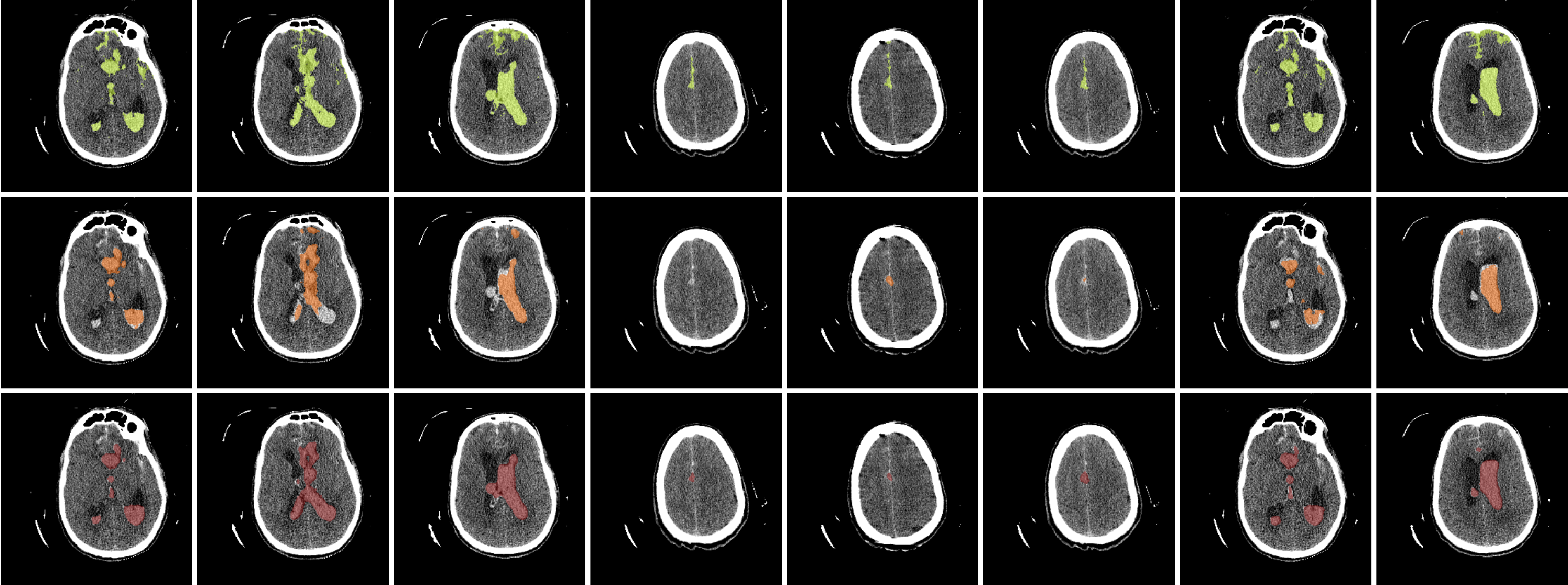}

  \caption{Segmentation results on few slices of the CQ 500 .The green regions($1^{st}\& 4^{th}\  row$) corresponds to the {\color{black}ground truth}, the orange regions($2^{nd}\& 5^{th}\  row$) shows segmentation {\color{black} without recursive strategy} and the red regions ($3^{rd}\& 6^{th}\  row$) shows segmentation after the {\color{black} after recursion}.}
  \label{fig:seg_CQ500_good}
\end{figure}
\begin{figure}[ht]
  \centering

       \includegraphics[ height=0.30\textheight,width=1\textwidth]{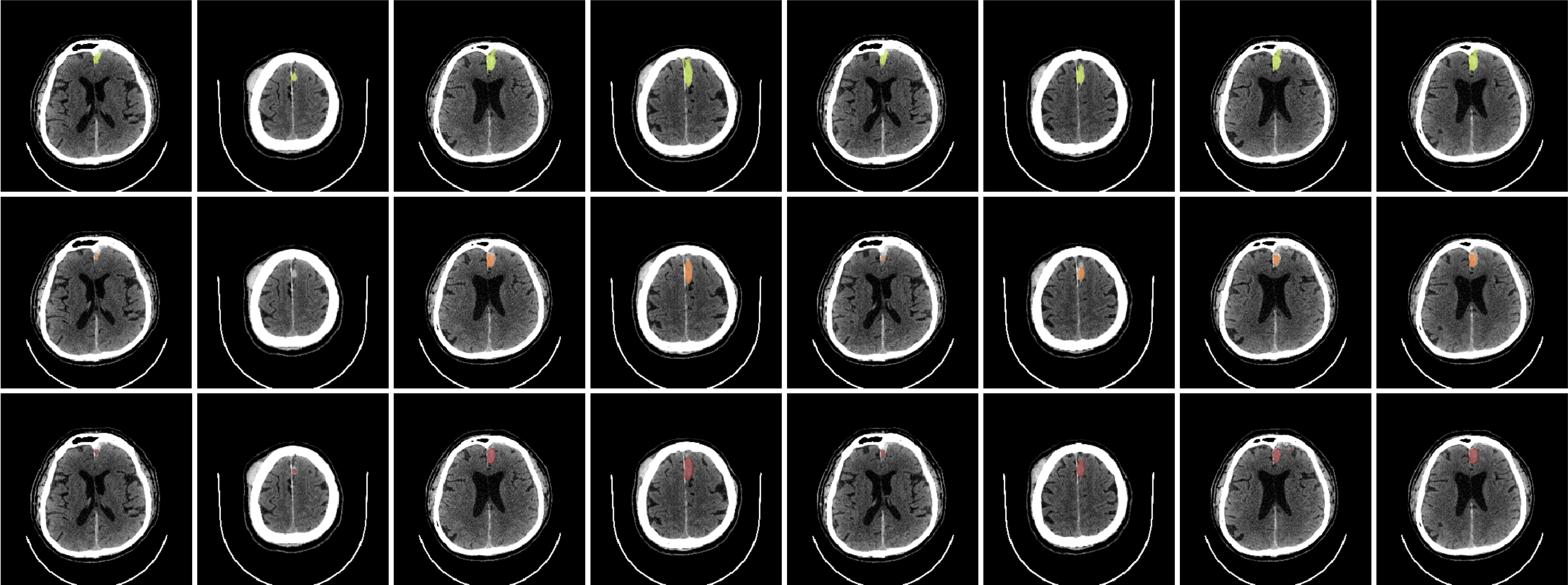}
     
     \vfill
     \includegraphics[ height=0.30\textheight,width=1\textwidth]{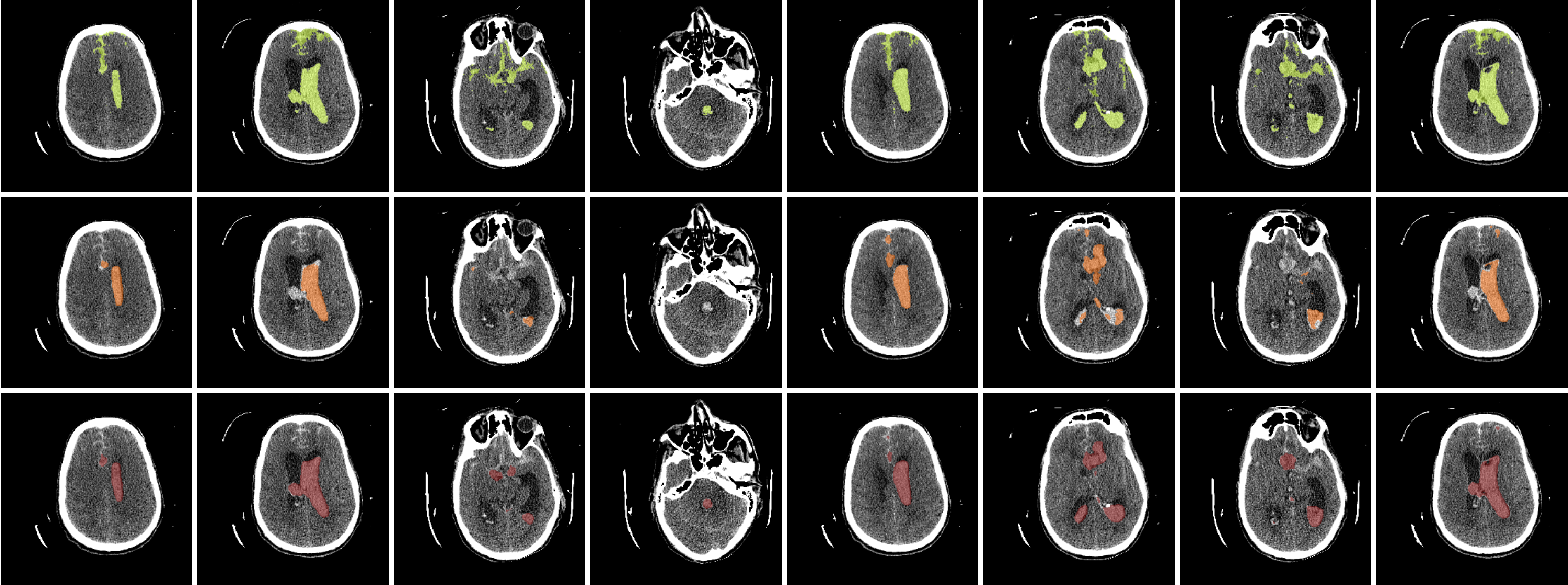}

  \caption{Segmentation results on few slices of the CQ 500 where the recursion helps identifying new regions for segmentation which were not identified without the recursive strategy. The green regions ($1^{st}~\&~ 4^{th}$ row) correspond to the {\color{black}ground truth}, the orange region ($2^{nd}~\&~ 5^{th}$ row) show segmentations {\color{black} without recursive strategy} and the red regions ($3^{rd}~\&~ 6^{th}$ row) show segmentation after {\color{black} recursion}.}
  \label{fig:seg_CQ500_new}
\end{figure}

\begin{figure}[ht]
  \centering
      \includegraphics[ height=0.30\textheight,width=1\textwidth]{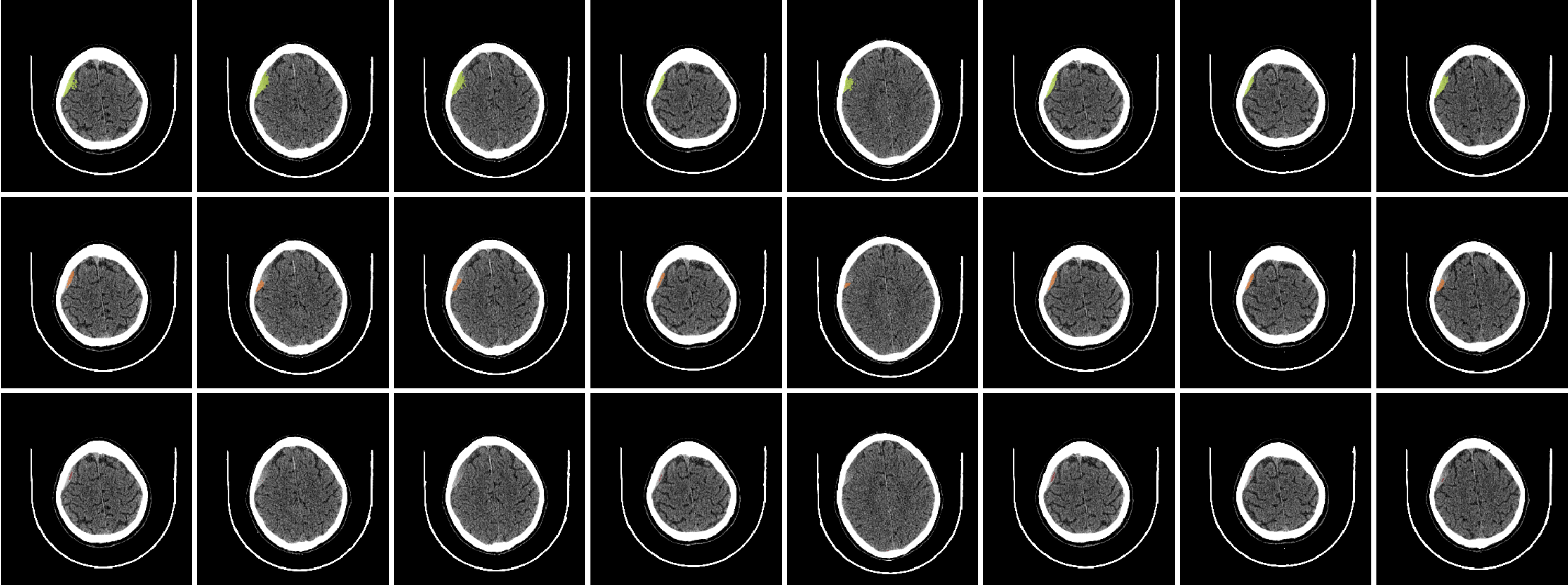}
  \caption{Some of the slices of the CQ 500 where a decrease in the dice score is seen. The green regions($1^{st}\ row$) corresponds to the {\color{black}ground truth}, the orange regions($2^{nd} \ row$) shows segmentation {\color{black} without recursive strategy} and the red regions ($3^{rd}\ row$) shows segmentation after the {\color{black} after recursion}.}
  \label{fig:seg_CQ500_bad}
\end{figure}

\clearpage

\subsection*{Result}\label{chap:result}
This sections shows the detailed comparison of the boxplot for all the available dataset. 
\begin{figure}[ht]
  \centering
  \subfloat[ \label{subfig:tum_before}]{%
       \includegraphics[height=0.2\textheight,width=0.48\textwidth]{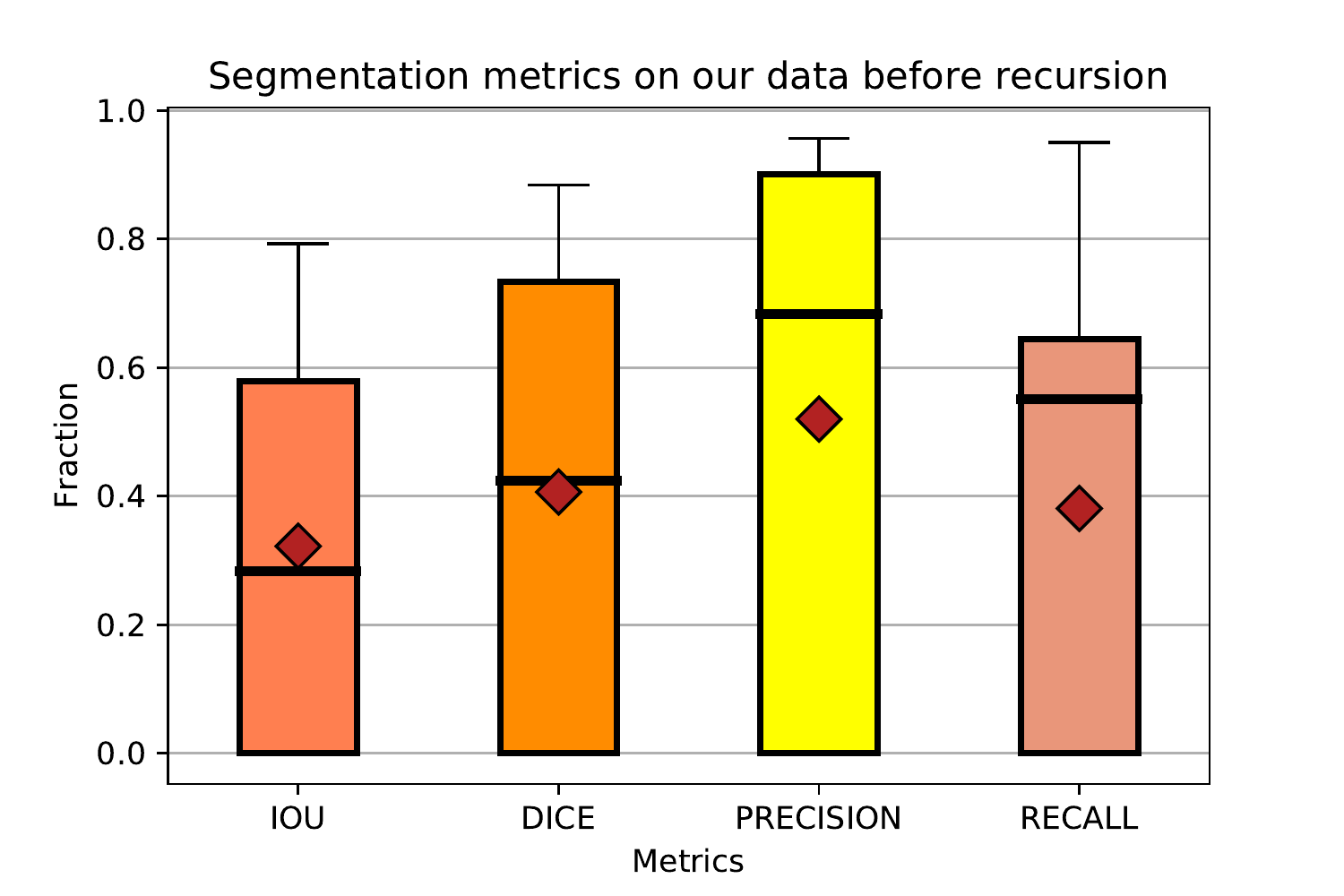}
     }
     \hfill
     \subfloat[  \label{subfig:tum_after}]{%
       \includegraphics[height=0.2\textheight,width=0.48\textwidth]{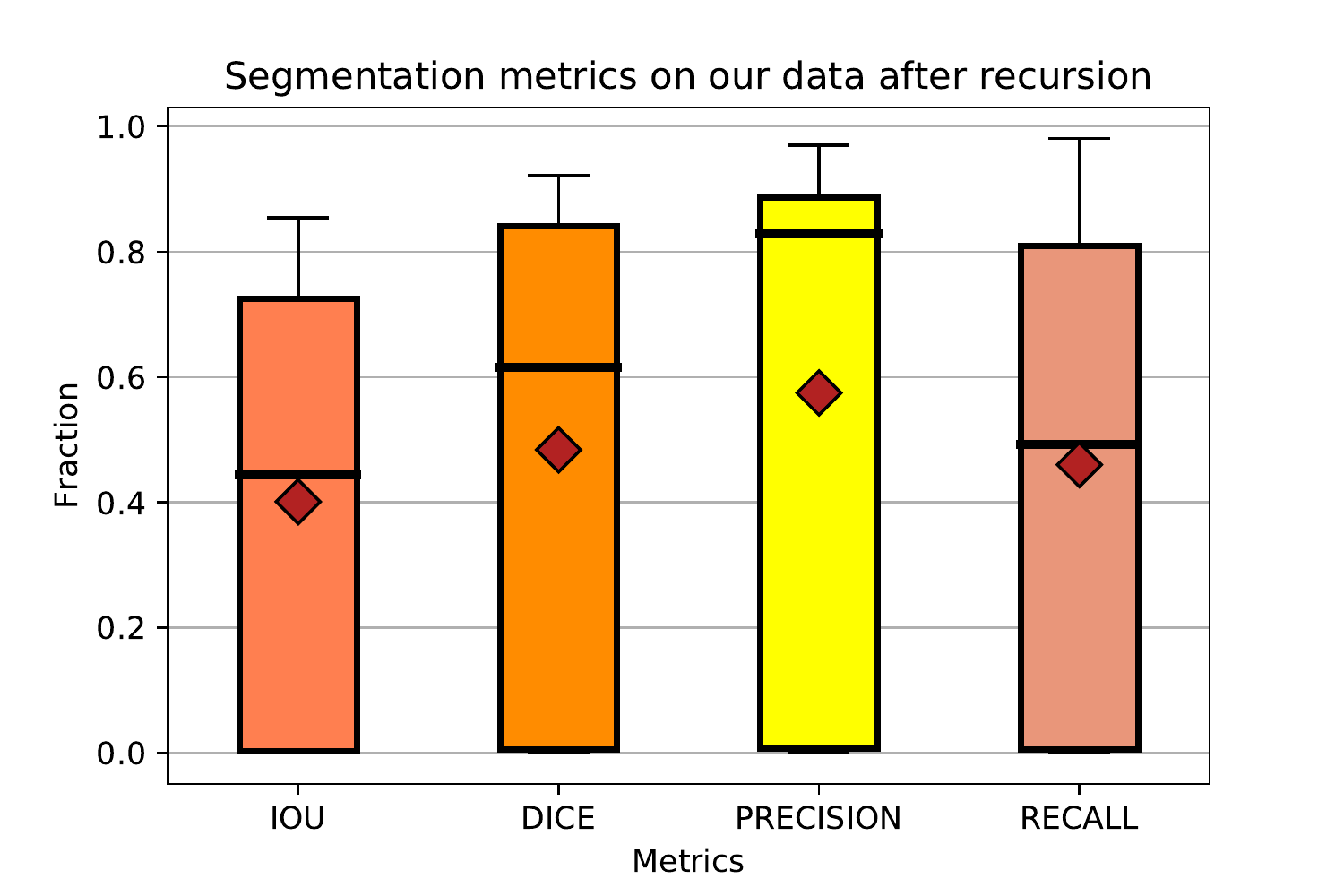}
     }
     \vfill
     \subfloat[ \label{subfig:cq_before}]{%
       \includegraphics[height=0.2\textheight,width=0.48\textwidth]{images/CQ.pdf}
     }
     \hfill
     \subfloat[   \label{subfig:cq_after}]{%
       \includegraphics[height=0.2\textheight,width=0.48\textwidth]{images/CQ_rec.pdf}
     }
  \vfill
     \subfloat[ \label{subfig:physio_before}]{%
       \includegraphics[height=0.2\textheight,width=0.48\textwidth]{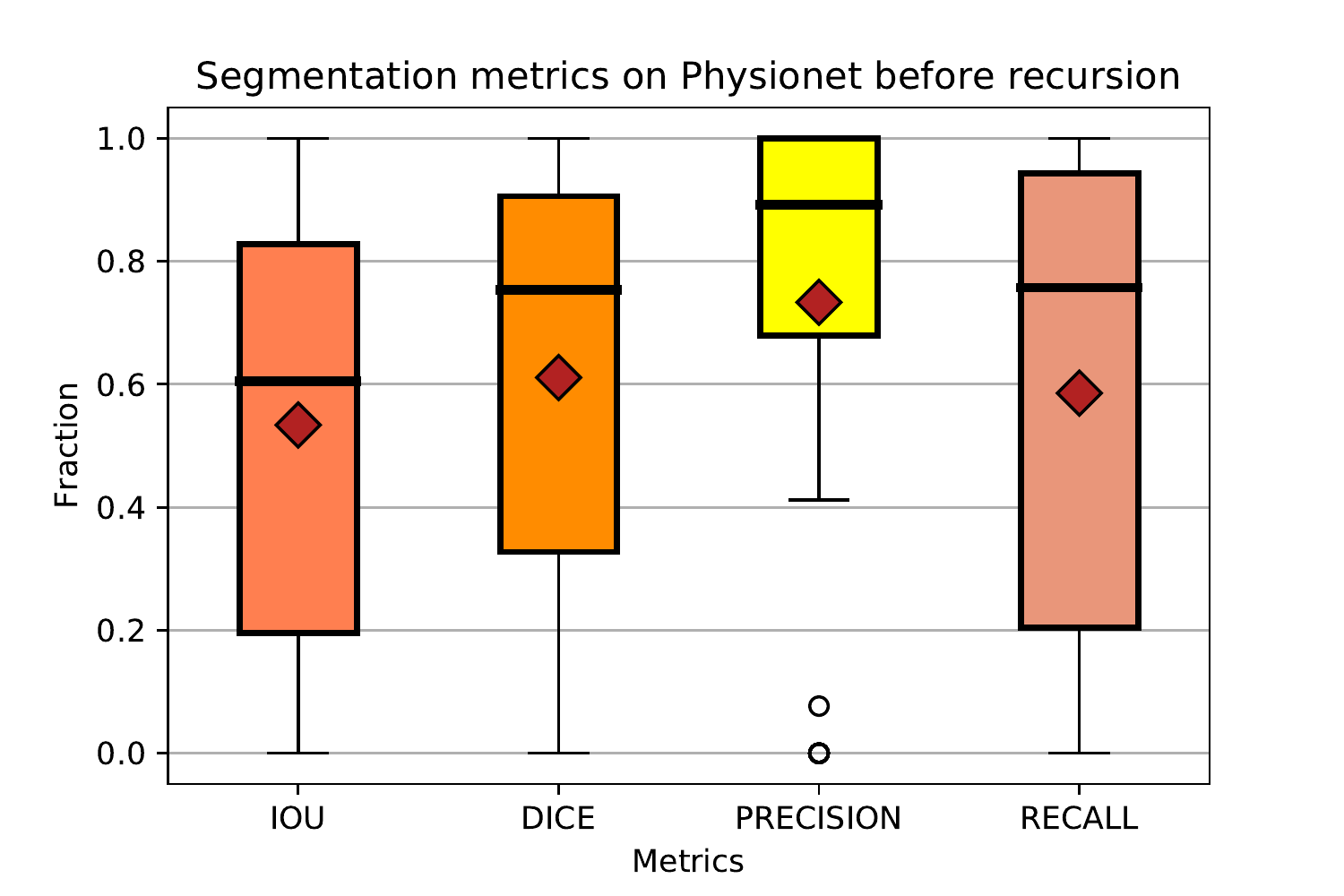}
     }
     \hfill
     \subfloat[   \label{subfig:l2}]{%
       \includegraphics[height=0.2\textheight,width=0.48\textwidth]{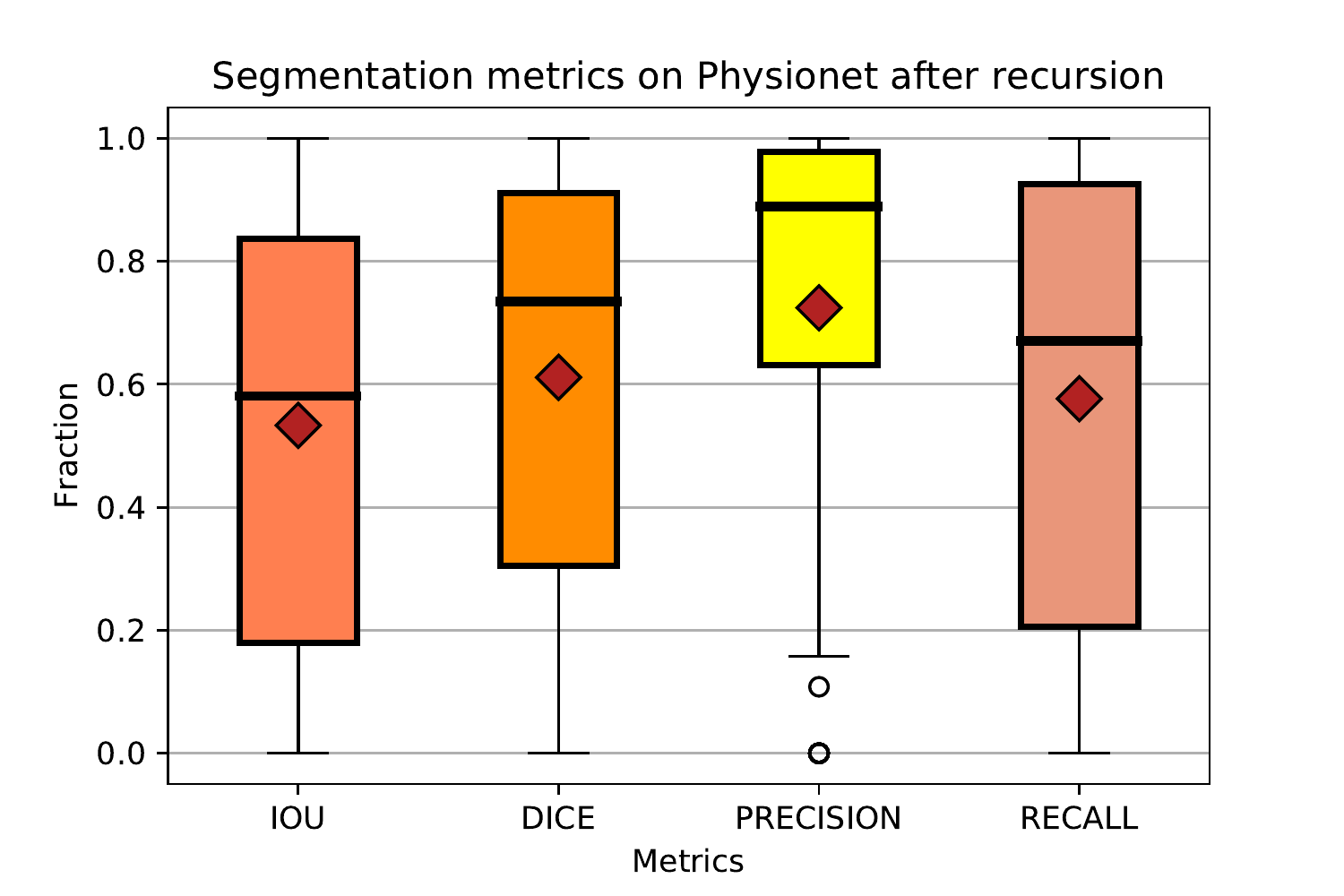}
     }

  \caption{Boxplot for segmentation metrics for all available datasets. }
  \label{fig:boxplot}
\end{figure}

\end{document}